# GPT-Micro: A Large Language Paradigm for Accelerated, Inexpensive, and Thermodynamics-consistent Discovery of Constitutive Models in Manufacturing

Soumik Dutta[a], Kiarash Naghavi Khanghah[b], Sania Shree[a], Logan McNeil[c], Thomas Feldhausen[d,e], Hongyi Xu[b*], Rajiv Malhotra[a*]

[a]Department of Mechanical and Aerospace Engineering, Rutgers University, USA

[b]Department of Mechanical, Aerospace & Manufacturing Engineering, University of Connecticut, USA

[c]Edison Welding Institute, USA

[d]Manufacturing Science Division, Oak Ridge National Laboratory, USA

[e]Department of Aerospace and Mechanical Engineering, University of Texas at El Paso, USA

[*]Email: rajiv.malhotra@rutgers.edu

*Email: hongyi.3.xu@uconn.edu

## Abstract

Constitutive modeling of the relationship between process-imposed material states and fundamental material properties is critical to control of material microstructure in manufacturing processes. The limited accuracy resulting from the typical reliance on fallible human expertise and intuition for postulation and revision of the model's functional form results in incremental and time-consuming model discovery. Conventional Machine Learning (ML) incurs significant cost and time of data generation. Model discovery using Large Language Models (LLMs) suffers from the above issues and/or ignores the inviolability of fundamental thermodynamics laws. This work creates a novel GPT-Micro paradigm for autonomous, data sparse, and thermodynamics-compliant discovery of de-novo constitutive models. This framework seamlessly integrates semantic knowledge extraction from literature, enforcement of thermodynamics-based conservation laws, and sparse datasets, with LLM-driven generation and refinement of model hypotheses. Validation is performed for a long-intractable constitutive modeling problem in a printed electronics process testbed. This reveals significant and simultaneous advantages over the state-of-the-art including: (a) More than 70% reduction in data burden relative to ML-based modeling without loss in accuracy; (b) 400X reduction in model discovery time after data generation, from months to hours, relative to human-driven modeling; (c) Discovery of models with novel functional forms without subjective human choice of a starting hypothesis; (d) Enhanced physics-rooted trustworthiness, human interpretability, and mechanistic insight via synthesis of compact, conservation-compliant, and physically complete analytical models. The potential of GPT-Micro to realize rapid, low-cost, physically trustworthy, and interpretable microstructure modeling across the manufacturing landscape is discussed.

## 1. Introduction

### 1.1. Importance of constitutive modeling

Computational modeling of microstructure evolution in manufacturing processes is critical for design and control of the fabricated part's material properties due to the significant cost and time of experimental trial-and-error. Such models quantitatively relate process-imposed material state (e.g., temperature or strain) to microstructure descriptors (e.g., grain or void size) for scalability across geometries and process parameters. For example, computational prediction of geometry-dependent surface roughness and grain structure in Additive Manufacturing can be used to design temporal variation of key process parameters[1, 2], and computational-prediction-driven microstructure control has driven cost reduction and quality improvement for cast parts[3, 4]. Figure 1 shows the key components of such computational models.

The first component is the thermodynamics-based laws of conservation of mass, momentum, and energy (Fig. 1a). The second component is the constitutive model that relates microstructure-relevant material properties to process-induced material states (Fig. 1b), e.g., relating diffusion coefficients and rate constants for phase composition (material properties) during Additive Manufacturing to temperature or strain (material state). The third component is the numerical solution method that solves the combination of conservation laws and constitutive laws over a desired spatiotemporal domain (Fig. 1c). Note that using constitutive models ensures physical consistency via incorporation of conservation laws, in contrast to purely-correlation-based direct modeling of process parameter-microstructure relationships.

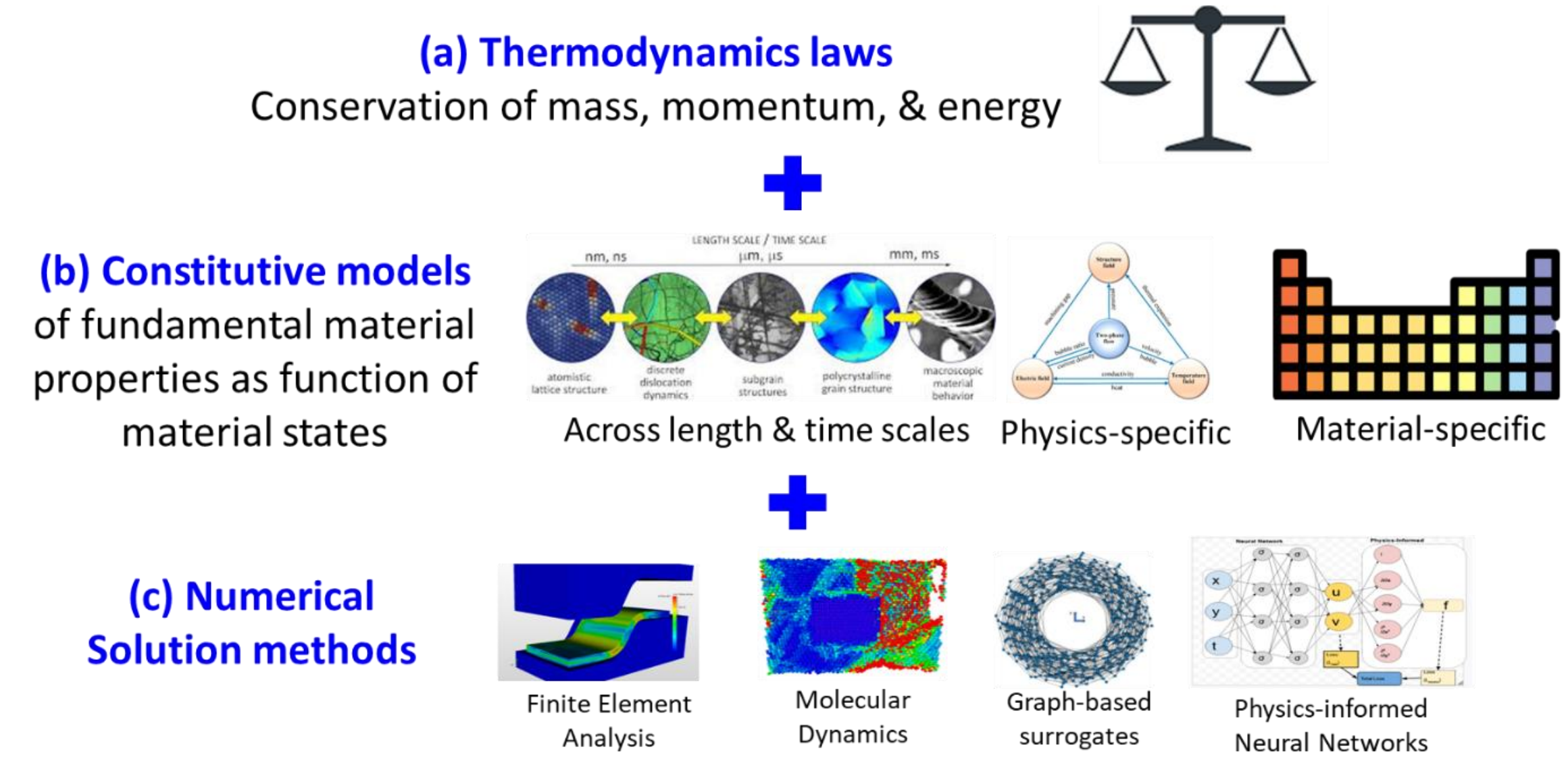

Fig. 1: Components of computational models for microstructure prediction and control.[5-7]

Significant effort has been expended on solution methods. Such methods include direct numerical simulations such as Finite Element Analysis (FEA[8, 9]), Molecular Dynamics[10], Phase-Field Modeling[11], and Cellular Automata[12]; and Machine Learning techniques like Physics Informed Neural Networks (PINNs)[5], Deep Operator Neural Nets (DeepONets)[6], and Graph methods[7]. Further, conservation laws are universally defined and invariant. But constitutive models are specific to the material and to the process-imposed physics and material states, e.g., combined effects of stress and temperature in sintering on grain size and porosity[13, 14]. They also have to incorporate microstructural attributes across multiple length and time scales, e.g., from sub-grain voids to cross-grain phase composition in additive manufacturing. Thus, constitutive models are often high dimensional by nature and are material- and process-specific.

The first step in the discovery of a constitutive model is identification of its functional form, i.e., the mathematical structure of how the material state descriptors are combined to determine material properties. The second step is calibration of this postulated form's material-specific coefficients. These steps are revised till microstructure predictions from solving the combination of conservation laws and constitutive models are sufficiently close to the ground truth microstructure data.

The above restriction on conformance to ground truth data within the confines of thermodynamics laws makes constitutive modeling more trustworthy than direct regression of process parameters or material states to microstructural attributes. Practically, discovery of novel functional forms of constitutive models is often necessary for new materials, e.g., for novel high entropy alloys in additive manufacturing[15]; for new processes that impose unusual material states, e.g., hybrid magnetic-additive processes[16]; and when insight is needed to guide innovation, e.g., understanding the impact of pressure on voids and grain structure has motivated hybridization of metal forming with additive manufacturing[17].

### 1.2. State-of-the-art in constitutive modeling

Based on the above discussion, the discovery of accurate conservation-law-compliant constitutive models is a keystone for microstructure modeling. Several optimization methods are available to calibrate the numerical coefficients of a known functional form of a constitutive model. Thus, the problem is to discover an appropriate functional form of the constitutive model.

The state-of-the-art suffers from a tradeoff between model scalability and data burden, as shown in Fig. 2. Model scalability refers to both prediction accuracy and the ability to holistically incorporate a large and diverse set of material states and properties. The data burden refers to the required amount of high-fidelity data from experiments and/or simulations. Methods based on in-situ sensing [18] are not discussed here, since such methods focus on quality control and do not provide predictive models that can be used for a-priori process design.

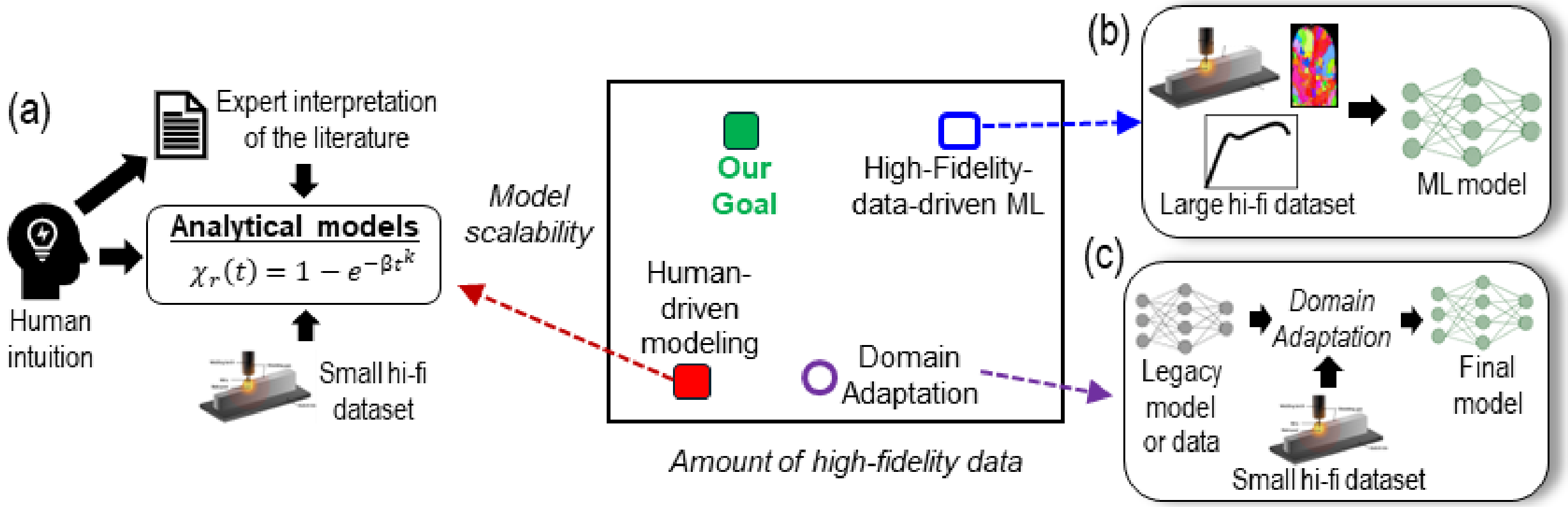

Fig. 2: Schematic of scalability-data tradeoff in the state-of-the-art constitutive modeling methods.

The major approaches for constitutive modeling in the state of the art are as follows:

*(1) Human-driven modeling*: The traditional human-driven approach synthesizes a model form based on intuition, expert interpretation of the literature, and a small amount of high-fidelity data (Fig. 2a). This dependence on human intuition/expertise for postulation of the model's functional form, often across a high-dimensional space of material properties and microstructure attributes, limits the model's prediction accuracy. As a result, incremental efforts by multiple researchers over long spans of time are required to incrementally develop sufficiently accurate constitutive models.

For example, while incremental sheet metal forming saw renewed interest in 2005 it took more than a decade to create accurate human-discovered constitutive models of metal fracture during the process[19]. Similarly, human-discovered constitutive models of phase composition and void size during electrically-assisted incremental forming are still unavailable more than a decade after process demonstration[20]. Despite experimental investigations and microscale modeling of the binder jet printing process for more than a decade, human-driven constitutive models cannot capture part-scale evolution of grain size, voids, cracking, and surface finish.[21] Despite advances in microscale[22] and bead-scale modeling[23, 24] in metal Additive Manufacturing there is a lack of human-discovered constitutive models that can predict geometry-dependent spatial heterogeneity in void size and surface finish due to post-solidification thermal cycling. Similarly, human-discovered constitutive models of phase evolution during metal Additive Manufacturing often yield inaccurate part-scale predictions since they require subjective human choice of the model's functional form for diffusional/diffusion-less phase transformations based on arbitrary thermal boundaries for isothermal, quasi-static, and non-isothermal behavior.[25]

Overall, human-driven constitutive modeling has limited scalability and is infeasibly slow due to its reliance on human expertise and intuition for postulation of the model's functional form.

*(2) High-Fidelity-data-driven Machine Learning*: The above limits of human-driven modeling have recently led to interest in Machine Learning (ML). ML methods like neural networks use high-fidelity data from experiments or simulations to correlate process-imposed material states to microstructural attributes (Fig. 2b).[26, 27] Human postulation of the model's functional form is not needed, unlike human-driven constitutive modeling and older data-driven methods like Surface Response Methodology.

But this approach incurs substantial time and cost associated with the significant number of high-fidelity experiments or simulations required to span the diverse combinations of material states and microstructure descriptors. For example, correlating complex spatiotemporally varying thermal cycles to multiple, multiscale, microstructure descriptors in metal Additive Manufacturing requires hundreds of thousands of dollars and months to years of experimentation despite the availability of parsimonious sampling methods like Active Learning.[28] Note that physics-informed surrogate models like PINNs and DeepONets suffer from the same dependence on human intuition and expertise as human-driven modeling.[5, 6, 29] This is because such surrogate models are either trained on data from physics-based models that need constitutive models or require constitutive models to be included in the loss functions.

Overall, high-fidelity-data-driven ML incurs significant time/cost burden to achieve model scalability.

*(3) Domain Adaptation*: Methods like Transfer Learning and Continual Learning achieve a middle ground between above techniques.[30-33] Specifically, Domain Adaptation modifies models or data from past work by updating it based on a small high-fidelity dataset to obtain the final ML-based constitutive model (Fig. 2c). Examples include models of curing for composites manufacturing[34], grain structure in Laser Powder Bed Fusion[35, 36], and surface structure in machining[37]. But the reliance on human expertise or intuition for subjective choice of legacy models/datasets that have similar data distribution as the local use-case yields unreliable accuracy that can deteriorate to infeasible levels.[30, 34]

The above scalability-data tradeoff slows down or prevents the industrial adoption of novel materials and manufacturing methods and control of part and material properties for enhanced product performance.

### 1.3. Goal of this paper

This paper aims to disrupt the above scalability-data tradeoff by enabling simultaneously autonomous, accelerated and data-thrifty discovery of novel constitutive models. Our approach will autonomously integrate LLM-enabled knowledge synthesis from semantically similar legacy literature, thermodynamics-driven conservation laws as constraints, data-driven feedback, and LLM-enabled hypothesis generation and refinement, to discover analytical constitutive models that are human interpretable and mechanistically insightful. We coin this paradigm as GPT-Micro, where "GPT" denotes use of Large Language Models based on Generative Pre-trained Transformers and "Micro" denotes discovery of constitutive models that describe evolution of the material's microstructure.

The remainder of this paper is organized as follows. Section 2 describes the approach and methodological novelties of GPT-Micro, along with the process testbed and the corresponding adaptation of GPT-Micro. Section 3 describes the results in terms of the discovered constitutive models and the demonstrated advantages of GPT-Micro over the state-of-the-art. Section 4 summarizes the learnings from this work with an eye towards future work in the manufacturing, materials and design context.

## 2. Methods

### 2.1. Working principle of GPT-Micro

Consider the general form of a microstructure evolution model in Equation 1, wherein the microstructural attributes of interest $\boldsymbol{\mu}$ are related to relevant material states $\boldsymbol{S}$ and state-dependent material properties $\mathcal{P}(\boldsymbol{S})$ via functional $f$. This $f$ is defined by thermodynamics based conservation laws and is known in advance.

The material states $\boldsymbol{S}$ can be obtained from relevant numerical simulations, e.g., temperature or stress from FEA. For example, when $f$ is the Allen-Cahn/Ginzburg-Landau equation for grain size or the Cahn-Hilliard equation for phase composition then $\mathcal{P}$ is the microstructural mobility coefficient[38] and chemical mobility tensor respectively, which depend on material states like temperature. Similarly, if $f$ is the Helmholtz free energy for void growth then $\mathcal{P}$ is the continuum damage variable that depends on material states like stress and strain rate.

$$\boldsymbol{\mu} = \boldsymbol{f}(\mathcal{P}(\boldsymbol{S}), \boldsymbol{S}) \tag{1}$$

$$\boldsymbol{\mu} = \mathcal{L}(\boldsymbol{S}) \tag{2}$$

$$\boldsymbol{\mu} = \mathcal{M}(\boldsymbol{S}) \tag{3}$$

GPT-Micro aims to discover the unknown, material-specific, and process-physics-dependent analytical form of $\mathcal{P}(\boldsymbol{S})$, i.e., the constitutive model that relates process-induced material states to fundamental microstructure-relevant material properties. The broad strokes of the working principle of GPT-Micro are illustrated in Fig. 3 and are as follows. A user supplies context keywords on the process, material, microstructure attributes, and states, that are of interest (Fig. 3a-b). These keywords are used in publisher APIs and open-source search engines for automated download of a paper corpus from literature (Fig. 3c).

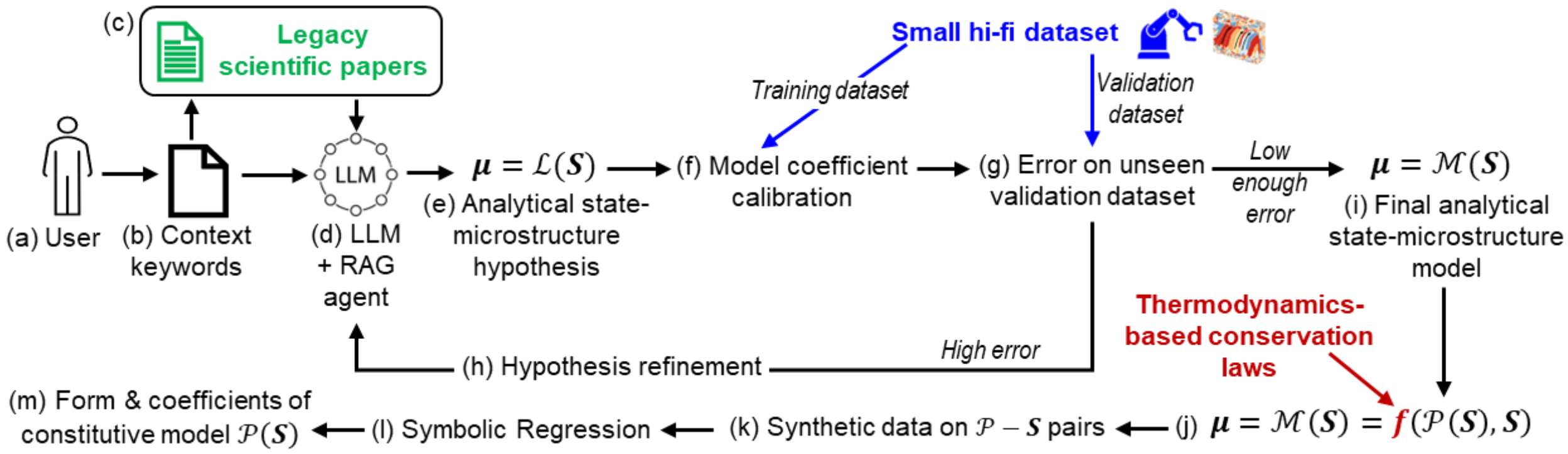

Fig. 3: Working principle of GPT-Micro.

A Large Language Modeling (LLM) agent with Retrieval Augmented Generation (RAG) is prompted to interpret text from the paper corpus and propose analytical hypothesis for the state-microstructure relationship (Fig. 3d-e). This hypothesis takes the form shown in Equation 2, wherein the hypothesized function $\mathcal{L}$ relates state $\boldsymbol{S}$ directly to microstructure attributes $\boldsymbol{\mu}$. The coefficients of function $\mathcal{L}$ are calibrated on a small high-fidelity training dataset (ground truth) using conventional nonlinear optimization (Fig. 3f).

If the resulting validation error on an unseen high-fidelity dataset is larger than a user-defined threshold (Fig. 3g) then the LLM agent is prompted to refine the form of hypothesis $\mathcal{L}$ (Fig. 3h) based on the hypothesis form and the corresponding validation error in the previous iteration. This iterative hypothesis refinement continues till an analytical state-microstructure model $\mathcal{M}$ with sufficiently low validation error is discovered (Equation 3 and Fig. 3i).

The above state-microstructure model $\mathcal{M}$ and the known conservation-law-defined form of the functional $f$ (i.e., equations 1 and 3) are combined to compute the values of $\mathcal{P}$ corresponding to diverse values of $\boldsymbol{S}$ (Fig. 3j-k). Conventional test-validation-test split of this synthetic dataset is used by standard Symbolic Regression methods[39] to discover the analytical expression and coefficients of $\mathcal{P}$ as a function of $\boldsymbol{S}$, i.e., the constitutive model (Fig. 3l-m).

The state-microstructure model $\mathcal{M}$ supplies Symbolic Regression with the larger dataset it requires without generating additional high-fidelity data. The discovered constitutive models are fed back into equation 1 to predict microstructure attributes as a function of material states. Section 2.4 describes additional details of GPT-Micro.

## 2.2. Key features of GPT-Micro

We discuss key characteristics of GPT-Micro in the context of the discussion in the introduction section.

1. *Discovery of the model's functional form*: GPT-Micro allows true model discovery. Specifically, it goes beyond calibration of coefficients of a model whose functional form is pre-decided by discovering both the functional form and the coefficients of the analytical constitutive model.
2. *Autonomous model discovery*: GPT-Micro is an autonomous approach to constitutive model discovery. The LLM agent is responsible for synthesis, interpretation, and adaptation of the literature into the functional form of the hypothesis followed by subsequent data-driven hypothesis refinement. Traditionally, this requires human expertise or intuition. Thus, GPT-Micro enables autonomous discovery of constitutive models without relying on fallible human expertise for formulation or choice of the model's functional form.
3. *Anthropomorphous model discovery*: GPT-Micro automates the traditional human-driven scientific method for model discovery. It iteratively proposes, calibrates, and refines hypotheses based on its interpretation of the literature, inviolable conservation laws and ground truth data. This makes GPT-Micro fundamentally anthropomorphic. This aspect is based on the insight that analytical equations are a language, i.e., variables representing physical quantities and standard mathematical operators constitute an alphabet and their composition into equations must align with reality. Since LLMs have significant ability to handle and manipulate language and interpret the literature, they are used here for postulation and refinement of the functional form of formal hypotheses.
4. *Reduced data burden*: As described in section 1.1, constitutive modeling requires identification of the functional form of the model and calibration of the model coefficients. Conventional ML-based approaches rely on high-fidelity data for both these steps. GPT-Micro leverages qualitative information on microstructure-state relationships from the downloaded paper corpus to reduce the amount of high-fidelity data required to identify the model's functional form. For example, past work on optically-driven sintering[40] can inform the functional form of how diffusion depends on time in electrically-driven sintering[41].
5. *Discovery of novel constitutive models from partial legacy information*: Every document in the downloaded paper corpus need not deal with the local use case. This is because our discovery of constitutive models that relate fundamental material properties to fundamental material states allows the LLM agent to compose information from diverse legacy sources that are related via partial mechanistic similarities. For example, qualitative discussions of the impact of solidification rate on grain structure during arc welding can guide the functional form of the corresponding state-microstructure model for Laser Powder or Direct Energy Deposition processes, given their partial mechanistic similarity in terms of deposition via melting and solidification. This contrasts with discovery of parameter-microstructure models, where papers must address the exact set of process, process parameters, and microstructural attributes as the local use case.[42, 43] Further, hypothesis refinement in GPT-Micro (Fig. 3h) allows ground truth data to modify the constitutive model's functional form for states or microstructure attributes with insufficient information in the literature.

   These aspects enable discovery of new constitutive models by cobbling together information from partially related sources in the literature and then filling any gaps using local ground truth data.
6. *Physics-rooted trust via thermodynamic consistency*: The use of conservation laws as an integral step for generation of synthetic data for Symbolic Regression (Fig. 3j-m) ensures that the discovered constitutive model is not just correlating input and output data (like the state-microstructure model in Fig. 3i) but is discovering constitutive models that are both accurate and thermodynamically consistent. This enables physically rooted trust in the constitutive model. This feature is possible with GTP-Micro because focusing on constitutive models as the goal for discovery allows a pathway to incorporate conservation laws while allowing microstructure prediction based on material states. On the other hand, direct correlation-based state-microstructure modeling can make predictions but cannot include the conservation laws.

7. *Human interpretability and mechanistic insight*: The closed-form analytical nature of the discovered constitutive models (i.e., $\mathcal{P}(\boldsymbol{S})$ in Fig. 3m) enables human interpretation and yields qualitative mechanistic insight. For example, a human process engineer can easily check how a specific material state affects a specific microstructure attribute, thus easing decision making on how to manipulate the state to control a specific microstructural attribute. Further, the closed-form nature of the discovered models lends itself to facile differentiation-based optimization for control.

To summarize, GPT-Micro integrates qualitative LLM-driven knowledge synthesis from the literature (even if partially related), constraints imposed by thermodynamics-driven conservation laws, small ground truth datasets, and LLM-enabled anthropomorphic hypothesis generation and refinement, to autonomously discover the functional form and coefficients of novel analytical constitutive models.

### 2.3. Methodological novelties

The autonomous nature of constitutive model discovery (point 2 in previous section) addresses the limited scalability of human-driven modeling. Specifically, the LLM agent eliminates dependence on the expertise and intuition of a human modeler for interpretation, synthesis, and adaptation of literature. Such autonomy also eliminates human choice of a legacy model or legacy datasets for Domain Adaptation. The incorporation of partially related qualitative information from the literature (Fig. 3b-d) reduces the amount of high-fidelity data needed as compared to data-driven ML. GTP-Micro also reduces the data burden for Symbolic Regression, which would require a significant amount of high-fidelity data if not for synthetic data generation from the LLM-discovered state microstructure model (Fig. 3j-k). Thus, GPT-Micro goes beyond the most popular constitutive modeling techniques discussed in Section 1.2.

We also compare GPT-Micro to other approaches that use LLMs for model discovery. Pretraining of LLMs for identification of suitable constitutive relationships cannot incorporate conservation laws.[44] Further, novel functional forms of constitutive models cannot be discovered due to lack of hypothesis refinement. LLMs for iterative refinement of a constitutive model's coeffcients[45, 46] cannot reduce the data burden and requires fallible human intuition/expertise to define the model's functional form, e.g., the prejudicial assumption of hyper-elasticity[46]. Other work uses knowledge extraction from the literature (via RAG) but cannot discover new models with accuracy due to the lack of hypothesis refinement.[47]

LLM with RAG-based knowledge retrieval has been used for parameter-microstructure modeling[42], which inherently prevents consideration of conservation laws, as discussed in point 6 in the previous section. Further, the lack of hypothesis refinement prevents discovery of new functional forms of constitutive models and reduces model discovery to coefficient calibration. Another effort use hypothesis refinement with LLM-RAG.[43] But the inability to include conservation laws remained as the focus was on parameter-microstructure models. Other works on LLM-enabled contextual querying do not yield constitutive models.[48, 49] LLM-based model discovery outside the manufacturing and materials domain also suffers from one or more of the above issues.[50-52]

This discussion highlights the methodological novelty of GPT-Micro, i.e., seamless integration of LLM-enabled knowledge extraction from literature, compliance with thermodynamics-based conservation laws, ground truth from sparse high-fidelity datasets, and LLM-driven hypothesis generation and refinement, for the first time.

### 2.4. Methodological Details

This section details key methodological elements of GPT-Micro, i.e., knowledge extraction from literature, hypothesis generation and refinement for state-microstructure models, and constitutive law discovery.

### 2.4.1. Extraction of legacy knowledge

RAG is used to retrieve knowledge relevant to the state-microstructure relationship from the downloaded paper corpus, i.e., the steps shown in Fig. 3b-c. Figure 4 illustrates this knowledge extraction process. The user-supplied material states and microstructural attributes are used to create a knowledge extraction prompt (Fig. 4a-c). This prompt, as shown in Table 1, retrieves and summarizes textual descriptions of the state-microstructure relationship. The phrases in < > in Table 1 are supplied by the user. The downloaded paper corpus is converted into smaller semantically meaningful text chunks using LlamaParse (Fig. 4d-e). These chunks and the knowledge extraction prompt are converted into vector embeddings using OpenAI's text-embedding-3-small model (Fig. 4f). Subsequent semantic similarity search between the documents embedding and the prompt embedding is used to retrieve the most relevant chunks from the literature (Fig. 4g). The top 15 chunks are re-ranked to extract information that is most relevant to the state-microstructure relationship.[53] These re-ranked chunks and the knowledge extraction prompt are used by an LLM (OpenAI's GPT-4o-mini) to qualitatively define the relationship between states and microstructural attributes (i.e., 'retrieved information', Fig. 4k). For example, 'retrieved information' might define if the dependence of grain size during sintering on peak temperature is positive, negative, linear, nonlinear, exponential, logarithmic, independent, or some other mathematical relationship.

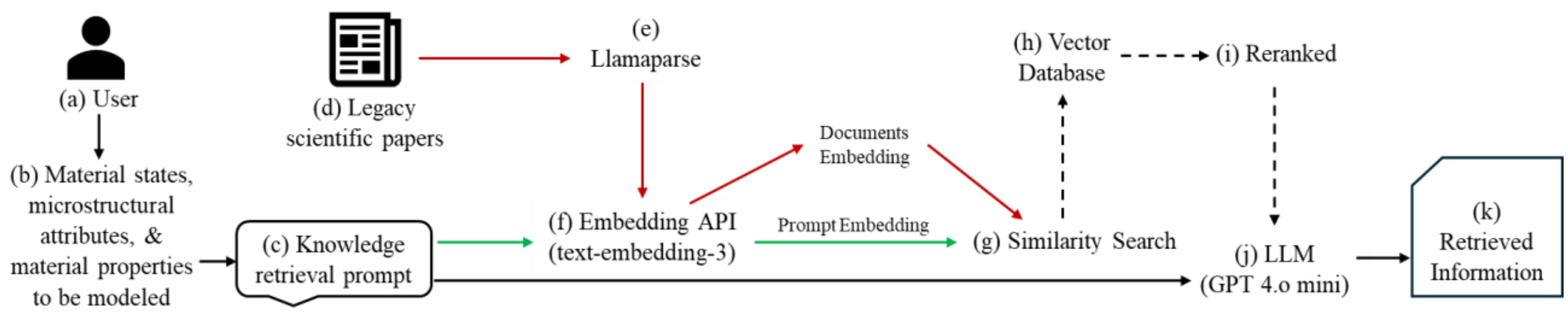

Fig. 4: Workflow for autonomous knowledge retrieval.

Table 1: Knowledge extraction prompt.

Input: PDF documents containing academic papers, *<material states>*, *<microstructure attributes>;*
Output: Qualitative information on relationship between *<material states>* and *<microstructure attributes>*

*Only use knowledge from PDFs. Retrieve and analyze direct relationships between <material states> and <microstructure attributes> as observed in academic papers. Identify how each of these <material states> influences the <microstructure attributes>. Specifically, examine whether the relationship is positive (increase in one variable leads to an increase in output variable), negative (increase in one variable leads to a decrease in the output variable, or follows a logarithmic, polynomial, or exponential or other mathematical trend. Provide concrete data or text from the academic papers that describe the behavior of the output variable in response to changes in input variables, including whether these changes are linear, nonlinear, logarithmic, or follow other patterns. Focus on how the output variable varies with respect to input variables.*

### 2.4.2. Hypothesis generation and refinement for state-microstructure model

The above-synthesized 'retrieved information' is used by an LLM to generate and refine the hypothesis for the state-microstructure model, i.e. the steps shown in Fig. 3e-h. Figure 5 illustrates this approach. An initial hypothesis generation prompt (shown in Table 2) instructs the LLM to use 'retrieved information' to generate 50 analytical equations (i.e., hypotheses) for the state-microstructure model (Fig. 5a-d). The LLM is prevented from relying on its pre-trained knowledge to avoid information contamination. Generating multiple initial hypotheses, with each hypothesis created independently from scratch, addresses potential hallucinations or inaccuracies during RAG-based knowledge extraction.[54] Gradient optimization is used to calibrate the coefficients of all the hypotheses by minimizing the Mean Square Error (MSE) on a small high-fidelity training dataset (Fig. 5e). The validation MSE is obtained for all the hypotheses on an unseen high-fidelity validation dataset (Fig. 5f). The top 20 hypotheses are selected and ranked according to their MSE (Fig. 5g). The hypothesis with the highest MSE for which the validation $R^2$ crosses a user-specified threshold is chosen as the final analytical state-microstructure model (Fig. 5h-j).

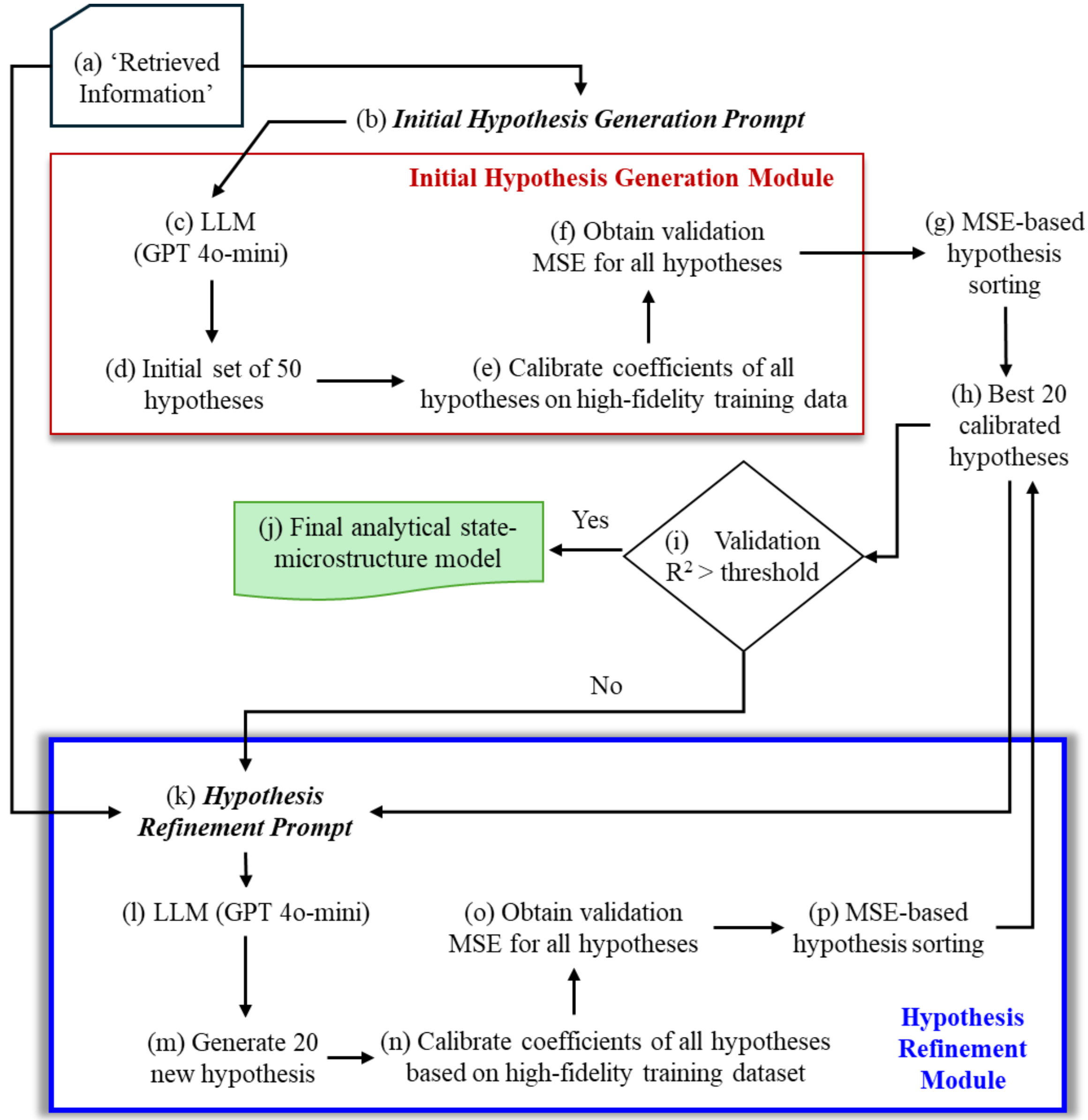

Fig. 5: Workflow for autonomous hypothesis generation and refinement.

The hypothesis refinement module is activated if the above $R^2$-based success criterion is not satisfied. The hypothesis refinement prompt in Table 3 is used by the LLM to generate 20 new hypotheses (Fig. 5k-m). This prompt instructs the LLM to access both the 'retrieved information' and the 20 hypothesis from the previous iteration along with their validation $R^2$ values (as 'summary'). It instructs the LLM to prioritize higher-performing prior models as the basis for generation of new hypothesis while exploring improvement strategies such as algebraic restructuring, term modification, or introduction of additional terms. The coefficients of all these hypotheses are calibrated, the corresponding validation MSE evaluated, and MSE-based hypothesis sorting performed, as described above (Fig. 5n-p). Identification of the best hypothesis whose $R^2$ value crosses the user-specified threshold yields the final analytical state-microstructure model (Fig. h-j), with the above hypothesis refinement continued if this success criterion is not met. The final model obtained (Fig. 5j) is tested on an unseen high-fidelity testing dataset.

Table 2: Initial hypothesis generation prompt

| Input: Qualitative nature of relationship between <material states> and <microstructure attributes> extracted from the literature (`Retrieved Information`); Output: Equations for output process variable as a function of input variables |
|---|
| *Based on the retrieved relationships between <material states>, affecting the <microstructure attributes> in the* `Retrieved Information`*, generate a Python function for the functions 'f(<microstructure attributes>)' used in the following equation. The equation is:*<br>*<material states> = f(<microstructure attributes>)*<br>*The function can be in various formats, including linear, exponential, logarithmic, or trigonometric functions.*<br>*The function should be formatted as:*<br>*def model (X, parameters, ...):*<br>    *input variables = X*<br>    *f = ...*<br>    *return f*<br>*Only utilize the retrieved information to generate the equation.* |

Table 3: Hypothesis refinement prompt

| Input: Qualitative nature of relationship between <material states> & <microstructure attributes> (`Retrieved Information`), top 20 hypothesis from the previous iteration and their calibrated coefficients and R2 scores (`Summary`), instructions to improve hypothesis performance, <material states>, <microstructure attributes>; Output: Refined equations for <microstructure attributes> as a function of <material states> |
|---|
| *Use the following information:* `Retrieved Information`.<br>*The results for the top 20 models were as follows in* `Summary`.<br>*The R2 scores indicate how well each model fits the data. An R2 score closer to 1 means a better fit, while a lower R2 score indicates a weaker fit. Therefore, the models with higher R2 scores and lower MSE are more reliable and should be given more consideration when creating new models.*<br>*Based on this information, please generate improved, complete function which lead to better R2 scores for modeling the <microstructure attributes> as a function of <material states>.*<br>*The model should be generated in this format:*<br>*def improved_model(X, parameters, ...):*<br>    *input variables = X*<br>    *f = ...*<br>    *return f* |

*Your task is to alter the equations based on previously retrieved information and previous best equations by combining or modifying operations (e.g., changing divisions to multiplications, combining terms, or introducing new terms) between <microstructure attributes> and <material states> that you retrieve from the above provided information.*

*The function must always conclude with return f. Please return the new Python functions in the same format as shown above. Focus on improving the R2 score of these models.*

### 2.4.3. Discovery of analytical constitutive models

The final analytical state-microstructure model is used to generate values of microstructural attributes ($\boldsymbol{\mu}$ in Equations 1 and 3) for diverse values of material states $\boldsymbol{S}$. The corresponding values for material properties $\mathcal{P}$ are computed from equation 1 by using the $\boldsymbol{\mu}$ and $\boldsymbol{S}$ and the conservation-law-defined functional $f$. This ensures that subsequent discovery of the constitutive model (i.e., $\mathcal{P}$-$\boldsymbol{S}$ relationship) is based on data that complies with conservation laws. The resulting $\mathcal{P}$-$\boldsymbol{S}$ dataset is fed to Symbolic Regression (PySR[39]) to discover an analytical expression of $\mathcal{P}$ in terms of $\boldsymbol{S}$ (i.e., the constitutive model). Note that Symbolic Regression simultaneously searches for both the equation structure and its associated coefficients.

## 2.5. Process testbed

This section describes the process and the modeling challenge for which GPT-Micro is validated.

### 2.5.1. Process description & challenge

The process testbed consists of printing-based deposition and thermal sintering of metal nanowires for fabrication of printed electronics circuits (Fig. 6a).[55, 56] This method fabricates planar[56], conformal[40, 57], and bulk-embedded electronics[58] with significant advantages in material window, throughput, geometric capability, and substrate compatibility. A key component of achieving targeted device performance is the ability to predict (and thus control) the degree of mass-transfer-driven shrinkage between nanowires during sintering. Molecular dynamics (MD) simulations have shown that large rotation at nanowire junctions nonlinearly regulates atomic diffusion and thus shrinkage (Fig. 6b).[59] This mechanism is absent in other nanoparticle shapes such as nanospheres or nanoflakes. The challenge is to develop models that relate the sintering-imposed temperature history to the inter-nanowire diffusion and rotation.

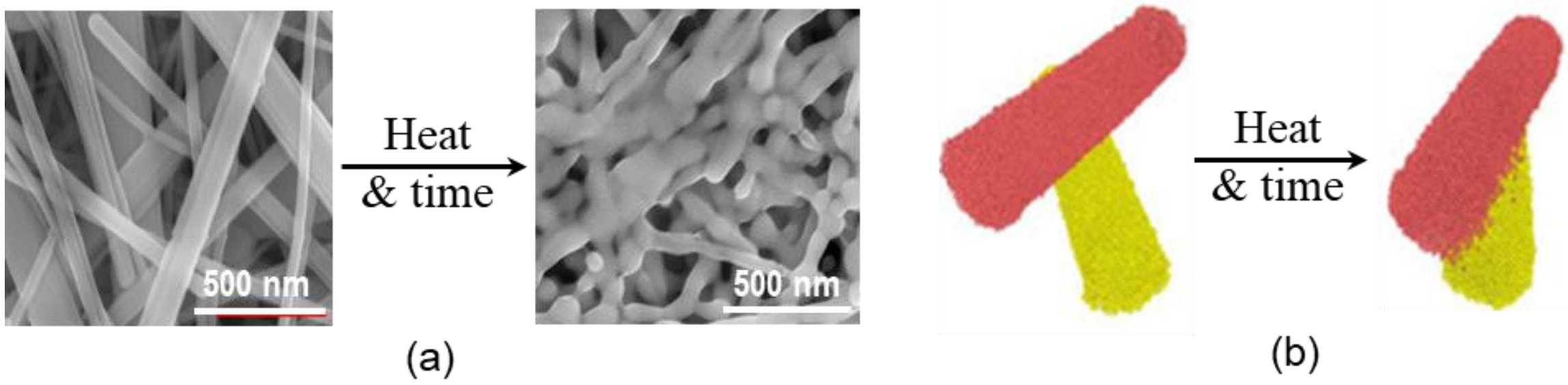

Fig. 6: (a) Experimental SEM micrographs of nanowire sintering process[56] (b) Molecular Dynamics simulations showing inter-nanowire rotation and shrinkage during the process.[59]

### 2.5.2. Mathematical description of the constitutive modeling problem

Conservation of mass, momentum, and energy can be used to express shrinkage $\delta$ and rotation $\theta$ at a nanowire junction as the differential equations in Equations 4 and 5.[60]

$$\frac{\partial \delta}{\partial t} = D_{eff}(R, T, \theta_0) \frac{4\gamma_s P(\delta, \theta)\Omega}{kT\sqrt{A(\delta, \theta)}} \left[\frac{2}{\delta} - \frac{P(\delta, \theta)}{A(\delta, \theta)} sin\,(\psi)\right] \tag{4}$$

$$\frac{\partial \theta}{\partial t} = \frac{\Gamma_{eff}(R, T, \theta_0)}{\eta \pi A^2(\delta, \theta)} \tag{5}$$

Here, $D_{\mathrm{eff}}$ is an effective atomic diffusion coefficient that represents cumulative effects of surface, grain boundary and volume diffusion. $\Gamma_{eff}$ is a fictitious rotational torque applied at the nanowire junction as the driver of inter-NW rotation. The nanowire radius is $R$, sintering temperature is $T$, and initial inter-nanowire orientation is $\theta_0$. In addition, $k$ is the Boltzmann constant, $\Omega$ is the atomic volume of the nanowire material (silver here), $\psi$ denotes the dihedral angle, $\gamma_s$ is the specific surface energy, and $\eta_c$ is the drag coefficient. $A$ is the area of the planar projection of the inter-nanowire neck and $P$ is the Ramanujan approximation of the 3D neck's perimeter, both of which are geometrically related to $\dot{\delta}$ and $\dot{\theta}$ via known functions.[60] Note that the dependence of $A$ and $P$ on $\delta$ and $\theta$ couples equations 4 and 5.

The constitutive modeling challenge is to use the high-fidelity data from MD simulations to identify the form and coefficients of analytical functions $D_{\mathrm{eff}}$ and $\Gamma_{eff}$(material properties) in terms of $R$, $T$, and $\theta_0$ (material states) so that $\dot{\delta}$ and $\dot{\theta}$ (microstructural attributes) can be predicted for circuit-scale predictions of sintering and electrical performance.[61] Note that Equations 4-5 state the conservation laws but do not assume the functional form of the constitutive models $D_{\mathrm{eff}}$ and $\Gamma_{eff}$ in any way,

### 2.5.3. Reasons for choosing this process testbed

Sintering of metal nanowires was experimentally reported in 2009.[62] But human-driven development of the above constitutive models has taken till 2021[61] despite significant efforts by multiple research groups. Thus, our process testbed is a good example of the limited accuracy and time-consuming nature of constitutive modeling methods that depend on human intuition/expertise (as discussed in Section 1.2). If GPT-Micro can autonomously discover accurate constitutive models in a shorter time, wherein this discovery time excludes the time needed for data generation, then this would quantify its ability to go beyond the above disadvantage of human-driven modeling. Reduction in the amount of high-fidelity MD data required as compared to conventional ML, without loss in accuracy, will quantitatively demonstrate reduction in data burden. For example, consider that the foregoing intuitive development of the constitutive models took $\approx$ 150 MD simulations corresponding to a total computational load of 27,500 CPU-days. Overall, this process testbed was chosen to quantify the envisioned advantages of GPT-Micro over the state-of-the-art.

### 2.5.4. Testbed-specific formulation of GPT-Micro

In the formalism of GPT-Micro (stated in Section 2.1), $D_{\mathrm{eff}}$ and $\Gamma_{eff}$ are the material properties $\mathcal{P}$; $T$, $R$, $\theta_0$ are the material states $\boldsymbol{S}$; $\dot{\delta}$ and $\dot{\theta}$ are the microstructure attributes $\boldsymbol{\mu}$, and equations 4 and 5 are the functional $f$. The legacy paper corpus was based on keywords listed in Supplementary Section S1. The high-fidelity dataset for discovery of the state-microstructure model was obtained from MD simulations. Equispaced sampling with five samples for each material state was employed with $R$ ranging from 7.5-12.5 nm, $T$ from 400-800 K, and $\theta_0$ from 0°-60°. MD simulations were performed to predict $\dot{\delta}$ and $\dot{\theta}$ for these unique combinations of material states. This dataset was split into 27 training samples, 27 validation samples, and 71 testing samples for discovery of the state-microstructure model. The higher number of testing samples relative to training and validation samples enabled robust validation of the discovered state-microstructure model. The user-defined threshold for the $R^2$ value in Fig. 5i was 0.98.

The discovered state-microstructure model and the conservations laws in Equations 4-5 were combined to discover the constitutive model (i.e., $\mathcal{P}$-$\boldsymbol{S}$ model). Distinct values of $R$, $T$, and $\theta_0$ were input to the state-microstructure model to compute $\dot{\delta}$ and $\dot{\theta}$, i.e., the left-hand side of equations 4-5. Forward Euler integration was used to compute the corresponding values of $\delta$ and $\theta$, i.e., the right-hand side of equations 4-5. The left-and and right-hand sides of equations 4-5 were equated to compute material properties $D_{\mathrm{eff}}$ and $\Gamma_{eff}$. Note that standard optimization can be used to compute the material properties for more complex forms of the conversation laws $f$. This computation yielded a synthetic dataset of $D_{\mathrm{eff}}$ and $\Gamma_{eff}$ for distinct combinations of $R$, $T$, and $\theta_0$. A total of 585 synthetic datapoints over 13 radius values, 9 temperature values, and 5 initial orientation values were generated with a split of 234 training points, 234 validation points and 117 testing points (all synthetic). This dataset was used by Symbolic Regression to compute the form and coefficients of the constitutive model, as illustrated in Fig. 3k-m. Supplementary Section S2 describes additional details of the LLMs and the Symbolic Regression.

**2.6. Assessment of advantages of GPT-Micro**

The advantage over high-fidelity data-driven ML was assessed by replacing LLM-driven discovery of the state-microstructure model with data-driven Feedforward Neural Networks (FNN), Support Vector Regression (SVR), Gaussian Process Regression (GPR), and Random Forest Regression (RF). These ML techniques are widely used in the literature for regression and range from compact representations (such as SVRs) to universal function approximators (such as neural networks). Grid search of hyperparameters was performed to obtain the best ML models. Supplementary Section S3 details these ML methods.

The reduction in data burden with GPT-Micro was quantified by examining the additional high-fidelity data required by the above ML techniques to achieve the same testing accuracy as GPT-Micro with 54 high-fidelity data points (27 training, 27 validation). For this purpose, the size of the high-fidelity dataset used by ML was increased to 343 (64 training, 64 validation, 215 testing) and 729 (125 training, 125 validation, 479 testing). The accuracy advantage of GPT-Micro for a low data budget was quantified by comparing the testing accuracy to ML methods, but for the same 54 high-fidelity data points for training and validation.

The time required by GPT-Micro to autonomously discover the constitutive model, after data generation, was compared to that needed for intuitive discovery by an expert since the first experimental inception of our process testbed. This quantifies how GPT-Micro can avoid the fallibility and the resulting time-consumption of human-driven modeling. Further, the form of the GPT-Micro-discovered models was compared to that of the human-discovered models to reveal the advantages of our approach in terms of mathematical compactness, human interpretability, and mechanistic insight.

**3. Results**

**3.1. Synthesis and form of discovered constitutive models**

The models synthesized by GPT-Micro are discussed for the smallest dataset, i.e., 54 training and validation samples. The final state-microstructure model is shown in equations 6 and 7, wherein $a_0$-$a_3$ and $b_0$-$b_6$ are material-specific model coefficients whose values are listed in Supplementary Section S4. The $R^2$ and normalized MSE values on the testing dataset (Fig. 7) indicate high prediction accuracy. Similar magnitudes of testing and validation error indicated good generalization, as is the case throughout this paper.

$$\dot{\delta} = a_0 \left(\frac{1}{R}\right)^{a_1} \frac{T^{a_2} t^{a_3}}{1+a_3\theta_0} \quad (6)$$

$$\dot{\theta} = b_0 \left(\frac{1}{R}\right)^{b_1} T^{b_2} log(t+1) + \frac{b_3 T\sqrt{\theta_0}}{R^{b_4}} + b_5\theta_0^2 t e^{-\frac{b_6}{R}} \quad (7)$$

Equations 8 and 9 show the analytical form of the constitutive models discovered from the foregoing state-microstructure model, i.e., for 54 training and validation samples.

$$D_{eff} = c_0 + \frac{c_1\sqrt{\theta_0}}{-T\theta_0 + e^{c_2 - sin(sinR)}} \quad (8)$$

$$\Gamma_{eff} = 10^{\left(\mathrm{d}_0 \mathrm{d}_1{}^{\theta_0} e^{\sin R} + \mathrm{d}_2(R + T^{\mathrm{d}_3}) - \mathrm{d}_4\right)} - \mathrm{d}_5 \quad (9)$$

The values of the material-specific model coefficients, i.e., $c_0$-$c_2$ and $d_0$-$d_5$ are listed in Supplementary Section S5. The corresponding $R^2$ and normalized MSE for $D_{eff}$ were 0.810 and 0.041 respectively, and for $\Gamma_{eff}$ were 0.952 and 0.049 respectively, indicating good fit with the synthetic dataset generated using the above state-microstructure models.

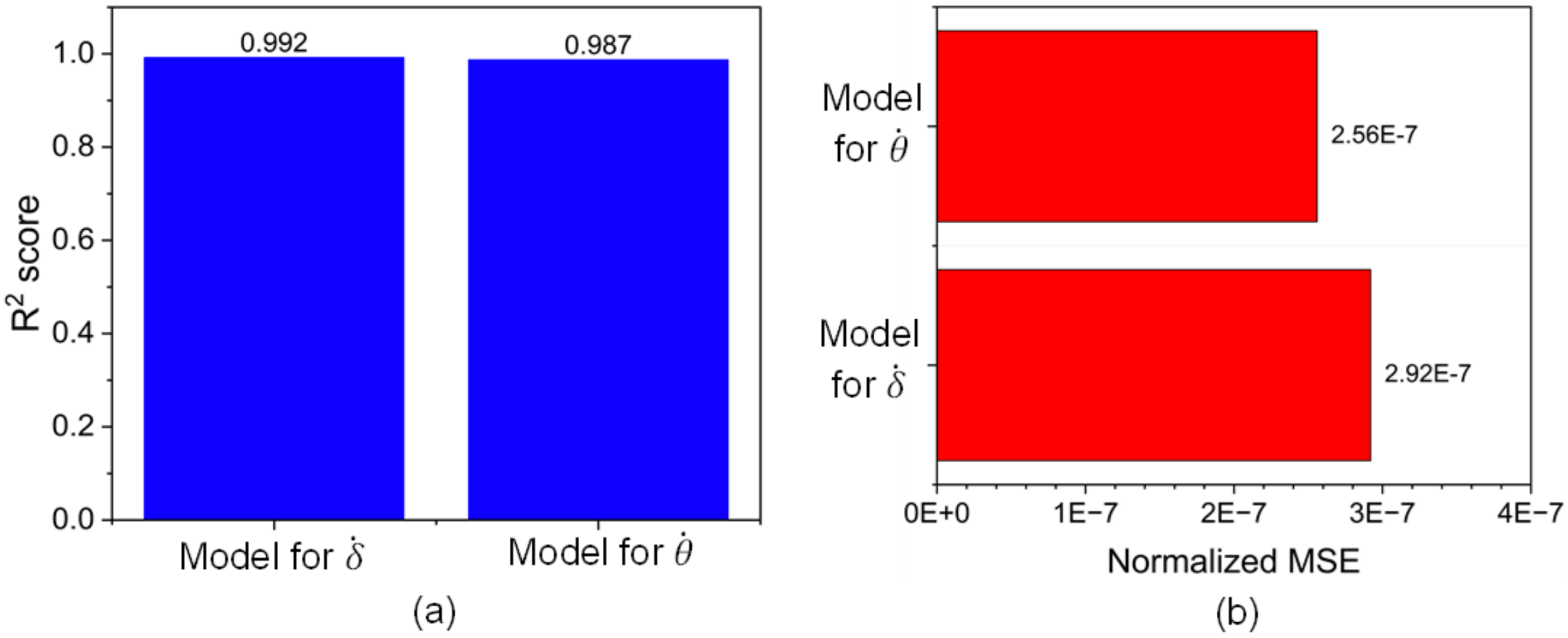

Fig. 7: (a) R2 score and (b) MSE for state-microstructure models.

Note that equations 4 and 5 are merely expressions of the constitutive laws and no human postulation of the form of state-microstructure models or constitutive models is required, unlike conventional human-driven modeling. Further, no human choice of legacy models or datasets is required, unlike Domain Adaptation techniques. Thus, GPT-Micro goes beyond the state-of-the-art constitutive modeling methods discussed in Section 1.2 via autonomous and anthropomorphic discovery of not just the model's coefficient values but also its functional form. These results support claims 1-3 in Section 2.2., i.e., GPT-Micro enables autonomously anthropomorphic discovery of the model's functional form and not just its coefficients.

### 3.2. Model discovery mechanism

Tracking the hypothesis accuracy during iterative discovery of the state-microstructure model, which is the necessary precursor to discovery of the constitutive model, yields interesting insight into the inner workings of GPT-Micro. The final state-microstructure model for $\dot{\delta}$ was obtained from the fifty initially generated hypotheses, i.e., without hypothesis refinement, in iteration 0 (Fig. 8). But Fig. 8 also shows that the initial hypotheses generation (i.e., iteration 0) did not yield a sufficiently accurate state-microstructure model for $\dot{\theta}$. In this case, three hypothesis refinement steps based on feedback from high fidelity data (via validation error) were necessary to improve the state-microstructure model for $\dot{\theta}$ to satisfactory $R^2$ values.

This difference between the discovery of models for $\dot{\delta}$ and $\dot{\theta}$ is better understood by examining the information on state-microstructure relationships that is autonomously extracted from the literature. The downloaded papers contained textual observations from experiments, theoretical models, and computational models of sintering of nanospheres and nanowires. But these papers did not contain any information on nanowire rotation during sintering, since recent work on MD simulations and theoretical modeling of such rotation[59, 60] was excluded from the downloaded paper corpus. This exclusion was deliberately done to reveal how GPT-Micro accommodates partially missing information on critical states or microstructural attributes, i.e., on inter-nanowire rotation here. The information extracted from the literature is shown in Tables S1 and S2 in Supplementary Section S6 and is discussed briefly below.

Table S1 indicates that the LLM extracts a physically correct functional dependence of $\dot{\delta}$ on material states. This is because the literature contains significant information on these qualitative dependencies for nanospheres and nanowires. For example, the LLM correctly assumes that nanowires with greater diameter will undergo slower shrinkage due to low surface-area-to-volume ratio. Thus, initial hypothesis generation based solely on information from literature correctly captures the qualitative dependence of $\dot{\delta}$ on material states and subsequent hypothesis refinement is not required.

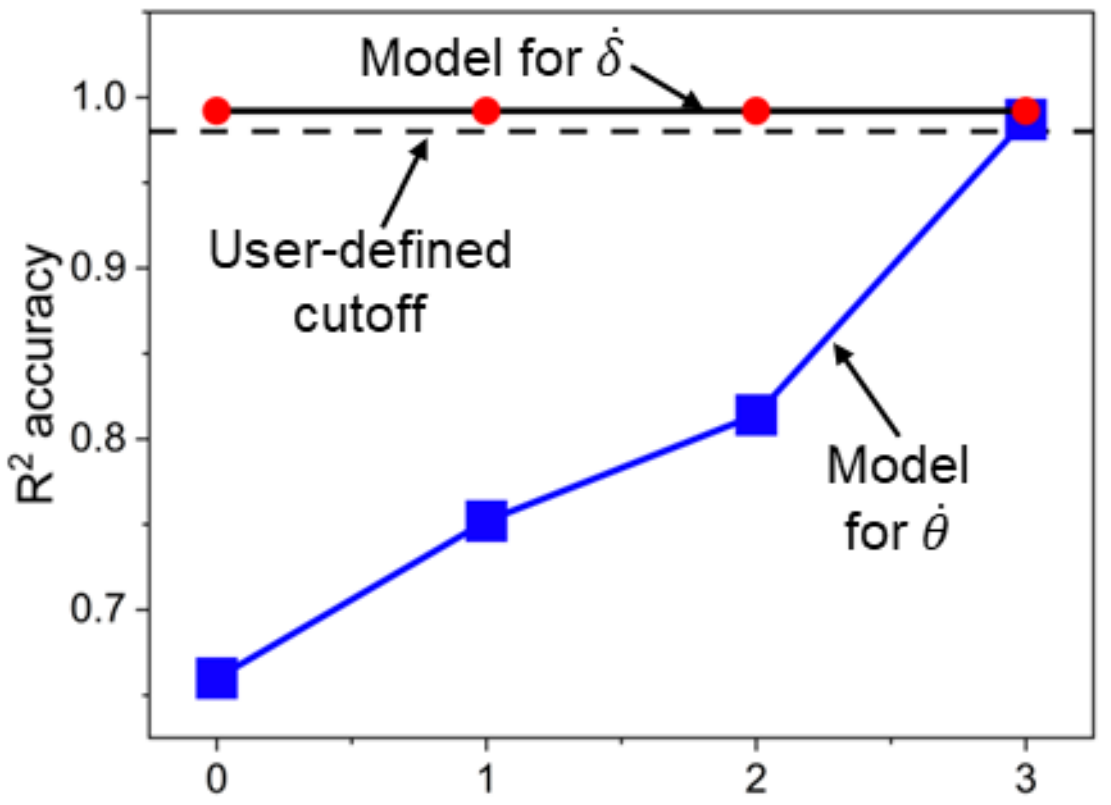

Fig. 8: $R^2$ score evolution with hypothesis refinement

Table S2 indicates that the LLM makes some unfounded and vague assumptions regarding the dependence of $\dot{\theta}$ on material states. For example, point 1 of Table S2 states that greater shrinkage (coalescence) due to greater atomic mobility increases the inter-nanowire rotation. This is a hallucination, given the lack of literature on inter-nanowire rotation in the paper corpus, and the author's manual observation that all the observations in the paper corpus (and indeed in the broader literature) on the partially related process of sintering of nanospheres indicate that shrinkage relieves rather than increases rotation. Thus, initial hypothesis generation based only on information extracted from the literature does not yield a viable model for $\dot{\theta}$. Instead, hypothesis refinement based on feedback from high-fidelity data is necessary (as in Fig. 8).

Thus, GPT-Micro discovers new state-microstructure models (and subsequently new constitutive models) by autonomously cobbling together whatever qualitative information is available in diverse partially related legacy sources (e.g., nanosphere sintering is partially related to nanowire sintering) and filling the gaps via iterative data-driven hypothesis refinement. This observation supports our claim 5 in Section 2.2.

### 3.3. Reduction in data burden

The data burden of constitutive model discovery equals that of state-microstructure model discovery, since the latter generates synthetic data for the former. Figure 9 compares the testing MSE and $R^2$ values for GPT-Micro with 54 training and validation datapoints to that achieved by the best performing conventional ML methods with 54, 128, and 250 training and validation points. Supplementary Section S7 shows the detailed accuracy metrics for conventional ML with each number of training and validation points.

Figures 9c-d show that GPT-Micro achieves more than 70% reduction in the amount of high-fidelity data required (i.e., along x axis) relative to conventional ML, to achieve the same normalized MSE. These plots also show that for the lowest data budget of 54 training and validation points GPT-Micro enhances prediction accuracy by more than 3X relative to conventional ML. These results support claim 4 in Section 2.2, i.e., the ability of GPT- Micro to significantly reduce the data burden relative to conventional data-driven ML without compromising prediction accuracy.

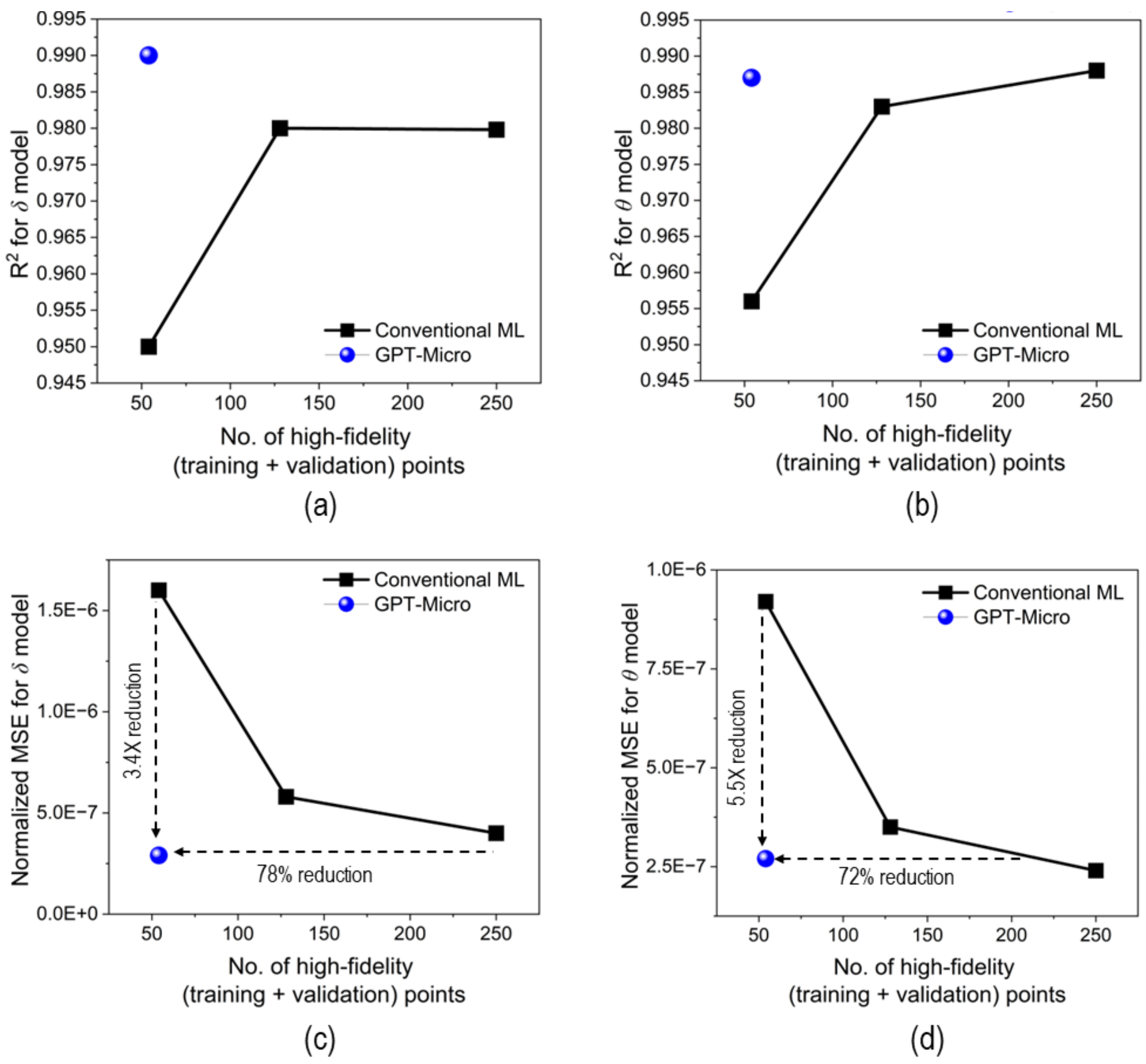

Fig. 9: Comparison of accuracy and data burden for GPT-Micro versus the best performing ML method.

### 3.4. Comparison to human-driven modeling and Domain Adaptation

Our process testbed was first reported experimentally in 2009.[62] The first constitutive model capable of capturing both atomic diffusion and inter-nanowire rotation was developed within the intuitive human-driven modeling paradigm by the lead author and his two PhD students in 2021[61], largely because intuitive consideration of coupled diffusion and rotation phenomena is intuitively difficult.

This human-driven modeling effort involved ≈ 150 MD simulations, of which 120 were used for functional formulation and calibration of the model (equivalent to model training), while the remaining 30 were used for testing the model on an unseen dataset. This amount of calibration data was found to be necessary to inform intuitive formulation of the model's functional form and optimization-based calibration of the resulting model coefficients with the composite effects of $R$, $T$, and $\theta_0$. Here, the time required for intuitive human-driven model discovery is quantified as that required to discover and calibrate the functional form of the constitutive model after all the required MD simulations are performed. For this testbed the human-driven modeling time was 6 months, primarily to iteratively assess the correct composition of functions that describe the individual and interactive effects of $R$, $T$, and $\theta_0$ on $D_{\text{eff}}$ and $\Gamma_{eff}$.

In contrast, GPT-micro required ≈ 10 hours on a 64-core CPU machine to discover accurate constitutive models after all the MD simulations (only 54 total for training and validation) were performed. This ≈ 400X reduction in the model discovery time (from months to hours) and the accuracy of the discovered constitutive models discussed in the previous section, highlights the ability of GPT-Micro to overcome the fallibility and the resulting time-consuming nature of human-driven modeling.

Note that GPT-Micro does not require subjective human choice of the initial hypothesis for the state-microstructure model, in contrast to Domain Adaptation, since LLM-RAG autonomously generates the initial hypothesis for subsequent refinement. Thus, the unreliability of Domain Adaptation due to the reliance on human choice of the initial hypothesis is inherently addressed by GPT-Micro.

### 3.5. Enhanced interpretability and mechanistic insight

The analytical nature of the GPT-Micro-discovered constitutive models in Equations 8-9 enables human interpretability and mechanistic insight. For example, Equations 8 and 9 show a complex exponential-double sinusoidal and base 10-exponential-sinusoidal functional dependence of $D_{eff}$ and $\Gamma_{eff}$ on nanowire radius $R$ respectively. Similarly, Equation 8 shows a linear dependence of $D_{eff}$ on temperature $T$ and Equation 9 implies a composite base 10-power law dependence of $\Gamma_{eff}$ on $T$. Such human-interpretable functional dependence enables easy analysis of the order of magnitude effects of material states on microstructure-relevant material properties. In addition to enabling qualitative decision-making for process control (e.g., increase or reduce nanowire radius), the analytical form of these GPT-Micro-discovered constitutive models allows gradient-based process optimization (e.g., of temperature) to achieve targeted microstructural attributes.

$$D_{eff} = \frac{e^{(f_0 R + f_1) + \left(\frac{f_2 R + f_3}{T}\right) + \left(\frac{-(\theta_0 - f_6)^2}{(f_4 T + f_5)^2}\right)}}{(f_4 T + f_5)} \tag{10}$$

$$\Gamma_{eff} = \frac{g_0 T^{g_1}\left(1 + \frac{g_4}{1 + \exp^{-g_5(\theta_0 - g_6)}}\right)}{R^{g_2} . T^{-g_3}} . e^{\left(\frac{-(\theta_0 - g_7)^2}{2 g_8{}^2}\right)} \tag{11}$$

The human-discovered models for $D_{eff}$ and $\Gamma_{eff}$ (Equations 10-11[61]) involved 18 and 24 mathematical operations respectively. The values of model coefficients for the human-discovered models are listed in our previously published work.[61] Equations 8-9 show that the GPT-Micro-discovered constitutive models require 40-50% fewer mathematical operations, i.e., 10 and 11 for $D_{eff}$ and $\Gamma_{eff}$ respectively. This reduced functional complexity of GPT-Micro-discovered models enhances interpretability and insight. For example, the GPT-Micro-discovered model in Equation 8 shows a clear and human-interpretable linear dependence of $D_{eff}$ on temperature since the corresponding $T$ term only appears in one function in this model. On the other hand, the appearance of $T$ in three different functions in the human-discovered model (Equation 10) confounds insight into the qualitative relationship between $D_{eff}$ and $T$. Similarly, the GPT-Micro-discovered model in Equation 9 shows a straigthforward and human-intepretable dependence of $log_{10}\Gamma_{eff}$ on an exponential form of $\theta_0$. But the appearance of $\theta_0$ in two different functions in the human-discovered model (Equation 11) confounds insight into the $\Gamma_{eff}$-$\theta_0$ relationship. Further, the GPT-Micro-discovered models require 50-75% fewer material-specific model coefficients than the human-discovered models.

Overall, GPT-Micro discovers more mathematically compact constitutive models that allow greater human interpretability and mechanistic insight for process control, which supports claim 7 in Section 2.2.

### 3.6. Physical consistency of discovered models

The combination of ML-discovered state-microstructure models with conservation laws to generate the synthetic dataset for subsequent Symbolic Regression yields insight into the thermodynamic consistency of conventional ML-based versus GPT-Micro-based model discovery. This analysis is performed for a training and validation set of 54 points, i.e., the smallest dataset used in this work, to incorporate a reduced data budget. The ML-discovered and GPT-Micro-discovered state-microstructure models were used to generate 585 equispaced synthetic data points over the $R$, $T$, $\theta_0$ space for subsequent Symbolic Regression, as discussed in Section 2.6.

For several instances across the above input space the solution of the ML-discovered state-microstructure models and conservation laws generated negative neck size and $D_{eff}$. This is thermodynamically inconsistent since it indicates a reduction in entropy, effectively violating the conservation laws. This observation is quantified in column 2 of Table 4 as the percentage of the input space for which physically consistent synthetic data can be generated, a value of 100% being obviously desirable. Only GPT-Micro-based and RFR-based state-microstructure models can generate physically consistent data over the entire input space. The challenge with ML-based modeling is further highlighted by the fact that FNNs, which are universal function approximators, do not generate any conservation-consistent synthetic data at all.

Table 4: Physical consistency and accuracy for ML-based vs. GPT-Micro-based constitutive models

| Method for state-μstructure modeling | Physically consistent synthetic data from conservation laws | Testing $R^2$ for $D_{eff}$ | Testing MSE for $D_{eff}$ | Testing $R^2$ for $\Gamma_{eff}$ | Testing MSE for $\Gamma_{eff}$ |
|---|---|---|---|---|---|
| GPT-Micro | 100% | 0.810 | 0.041 | 0.952 | 0.049 |
| FNN | 0% | - | - | - | - |
| SVR | 26% | - | - | - | - |
| GPR | 38% | - | - | - | - |
| RFR | 100% | 0.521 | 0.423 | 0.739 | 0.259 |

Further, the MSE for the RFR-based constitutive models was an order of magnitude greater than the GPT-Micro-discovered constitutive models (Table 4). This is explained by the fact that the RFR-discovered constitutive models (Equations 12 and 13) omit the critical effect of temperature $T$. Effectively, GPT-Micro discovers analytical constitutive models that are physically more complete than those discovered by ML. The material-specific model coefficients for equations 12 and 13 are listed in Supplementary Section S8.

$$D_{eff} = h_0 + h_1\left(\sqrt{\theta_0}exp\left(h_2 - sin(h_3) + h_4 sin\big(sin(R)\big)\right) - h_5\right) \quad (12)$$

$$\Gamma_{eff} = 10^{k_0 + k_1\left(k_2 {k_3}^{\theta_0} + sin(log(R))\right)} - k_4 \quad (13)$$

These results have the following implications. The direct use of ML-discovered state-microstructure models for microstructure predictions is not trustworthy since such models may be thermodynamically inconsistent over the entire space of material states (e.g., FNN-based model in Table 4). ML-based constitutive modeling can address this issue but may not be physically complete, e.g., the RFR method in Table 4. GPT-Micro-discovered constitutive models are both thermodynamically consistent and physically complete.

Overall, these observations support Claim 6 in Section 2.2 that GPT-Micro enables more trustworthy discovery of constitutive laws by assuring their physical consistency with thermodynamics-driven conservation laws and by enhancing complete encapsulation of the relevant physical phenomena.

## 4. Conclusions

This paper introduces a novel GPT-Micro paradigm for autonomous, inexpensive, and thermodynamics-compliant constitutive modeling in manufacturing processes. GPT-Micro is a unique autonomously anthropomorphic constitutive modeling framework that seamlessly integrates semantic knowledge extraction from literature, compliance with thermodynamics-driven constitutive laws, sparse ground truth datasets, and LLM-driven data-informed hypothesis generation and refinement. The following simultaneous advantages over the state-of-the-art are demonstrated:

1. GPT-Micro reduces the high-fidelity data required for constitutive model discovery by more than 70% as compared to traditional ML, without loss in accuracy (see Section 3.3).

2. GPT-Micro leverages whatever qualitative information is available in legacy literature and fills the gaps in it via iterative data-driven hypothesis refinement to discover novel functional forms of constitutive models, thus going beyond just calibration of model coefficients (see Sections 3.1-3.2).

3. The reduction in model formulation time from months to hours as compared to human-driven modeling highlights the ability to overcome the inherent fallibility and resulting time-consuming nature of intuition-based constitutive modeling (see Section 3.4).

3. Unlike Domain Adaptation methods, GPT-Micro does not require subjective human choice of the initial hypothesis based on a human expert's interpretation of the literature. Instead, it autonomously generates the initial hypothesis for subsequent refinement, as discussed in Section 3.4.

4. GPT-Micro discovers constitutive models that are more mathematically compact than human-discovered models, i.e., with 40-50% fewer mathematical operations and 50-75% fewer model coefficients, as discussed in Section 3.5. This enhances interpretability by human engineers, which is critical for qualitative and quantitative process control.

5. The constitutive models discovered by GPT-Micro are more trustworthy than ML-discovered state-microstructure models or constitutive models due to their assured consistency with thermodynamics-driven conservation laws (see Section 3.6). Further, GPT-Micro-discovered constitutive models are more complete than ML-discovered ones in terms of capturing relevant physical phenomena.

Overall, GPT-Micro goes beyond the scalability-data tradeoff that plagues traditional constitutive modeling and beyond the inability of emerging LLM-based process modeling efforts to simultaneously incorporate literature, ground truth data, conservation laws, and hypothesis generation and refinement.

GPT-Micro is easily scalable to other processes by reframing the constitutive laws and the definitions of the relevant material states, microstructural attributes, and material properties. For example, GPT-Micro could be used in the future to create constitutive models that describe the element-wise evolution of phase composition and grain structure descriptors as a function of element-wise temperature history in thermal FEA of Additive Manufacturing processes. We envision that GPT-Micro constitutes a scalable paradigm for systemically accelerated, inexpensive, and trustworthy modeling of microstructure evolution as a function of fundamental material states across diverse processes.

We believe that GPT-Micro constitutes a revolution in process modeling by eliminating the cons but preserving the pros of the state-of-the-art modeling techniques, thus going beyond both traditional human-driven modeling and the more recently emergent ML-based modeling. The demonstration of GPT-Micro for additional processes, and integration of ontological techniques to identify relevant states and microstructural attributes for a given application, are part of our ongoing work.

**CRediT authorship contribution statement:**

Soumik Dutta, Kiarash Naghavi Khanghah, Sania Shree: Data curation, Investigation, Methodology, Software, Validation, Visualization, Writing – original draft. Logan McNeil: Resources, Writing – review and editing. Thomas Feldhausen: Resources, Writing – review and editing. Hongyi Xu: Conceptualization, Funding acquisition, Project administration, Resources, Supervision, Writing – original draft, Writing – review and editing. Rajiv Malhotra: Conceptualization, Funding acquisition, Project administration, Resources, Supervision, Writing – original draft, Writing – review and editing.

**Acknowledgements**

This work was supported by the National Science Foundation, USA, grants 2414397, 2336448, 2434385, 2414398, the Rutgers Global grant, and the U.S. Department of Energy's Office of Energy Efficiency and Renewable Energy (EERE) under the Advanced Manufacturing Office Award Number DEEE0011029.